%% file: main.tex
\documentclass[conference, 9pt]{IEEEtran}
\usepackage{multirow}
\usepackage{verbatim}
\usepackage{graphicx}
\usepackage{amsmath, amssymb, amsthm}

\theoremstyle{definition}

\usepackage[capitalize]{cleveref}
\crefformat{equation}{(#2#1#3)}
\usepackage{enumitem}
\setlist[itemize]{noitemsep, topsep=0pt}
\usepackage{makecell}
\usepackage{siunitx}
\usepackage[caption=false, font=footnotesize]{subfig}
\usepackage[linesnumbered,ruled,lined]{algorithm2e}
\SetArgSty{textup}
\usepackage{lipsum}
\usepackage{tabularx}
\newcolumntype{Y}{>{\centering\arraybackslash}X}
\usepackage{cite}
\usepackage{color}
\usepackage{booktabs}
\usepackage{colortbl}
\usepackage{setspace}
\usepackage{amsmath}
\definecolor{mygray}{gray}{.9}
\usepackage[table,xcdraw]{xcolor}

\makeatletter
\IEEEtriggercmd{\reset@font\normalfont\fontsize{6.7pt}{7.2pt}\selectfont}
\makeatother
\IEEEtriggeratref{1}
\usepackage{xfrac}
\begin{document}
\bstctlcite{IEEEexample:BSTcontrol}
\setstretch{1}
\title{Learning Library Cell Representations in Vector Space}

\author{\IEEEauthorblockN{Rongjian Liang}
\IEEEauthorblockA{\textit{NVIDIA}\\
rliang@nvidia.com}
\and
\IEEEauthorblockN{Yi-Chen Lu}
\IEEEauthorblockA{
\textit{NVIDIA}\\
yilu@nvidia.com}
\and
\IEEEauthorblockN{Wen-Hao Liu}
\IEEEauthorblockA{
\textit{NVIDIA}\\
wenhliu@nvidia.com}
\and
\IEEEauthorblockN{Haoxing Ren}
\IEEEauthorblockA{
\textit{NVIDIA}\\
haoxingr@nvidia.com}
}

\maketitle
\thispagestyle{plain}
\pagestyle{plain}

\begin{abstract}

We propose \textit{Lib2Vec}, a novel self-supervised framework to efficiently learn meaningful vector representations of library cells, enabling ML models to capture essential cell semantics. The framework comprises three key components: (1) an automated method for generating regularity tests to quantitatively evaluate how well cell representations reflect inter-cell relationships; (2) a self-supervised learning scheme that systematically extracts training data from Liberty files, removing the need for costly labeling; and (3) an attention-based model architecture that accommodates various pin counts and enables the creation of property-specific cell and arc embeddings. Experimental results demonstrate that Lib2Vec effectively captures functional and electrical similarities. Moreover, linear algebraic operations on cell vectors reveal meaningful relationships, such as \textit{vector(BUF) - vector(INV) + vector(NAND)} approximating the vector of \textit{AND}, showcasing the framework’s nuanced representation capabilities. Lib2Vec also enhances downstream circuit learning applications, especially when labeled data is scarce.


\end{abstract}

\begin{IEEEkeywords}
representation learning, library cell, netlist optimization
\end{IEEEkeywords}

\input{tex/introduction.tex}

\input{tex/Preliminaries}
\input{tex/overview}
\input{tex/reguality_tests}

\input{tex/self_supervision}
\input{tex/model}
\input{tex/usage}

\input{tex/Experiment}
\input{tex/conclusion}




\vspace{-1mm}
\bibliographystyle{IEEEtran}
\bibliography{main.bbl}

\end{document}

%% file: tex/introduction.tex
\section{Introduction}


Library cell representations are vital for effective machine learning (ML)-based circuit analysis and optimization, as library cells are the fundamental building blocks of circuit netlists. Traditional methods often rely on manually defined features~\cite{barboza2019machine, xie2022preplacement, guo2022timing, zhang2020grannite}, requiring extensive expertise and feature engineering. Alternatively, one-hot encoding~\cite{fayyazi2019deep} demands large amounts of domain-specific training data, which may not always be available. To address these limitations, this work explores a data-efficient, self-supervised learning approach to generate cell representations.

Our method maps library cells into a continuous vector space, capturing semantic relationships and enabling ML models to operate in a simpler, structured space instead of the original high-dimensional, discrete cell space. Our work also aligns with the broader AI shift toward training foundation models~\cite{bommasani2021opportunities} on self-supervised data and adapting them to diverse downstream tasks. By providing a unified, self-supervised method for learning cell representations, this work has the potential to serve as a groundwork for developing circuit foundation models~\cite{chen2024large}.

While related efforts, such as DeepGate~\cite{li2022deepgate,shi2023deepgate2} and FCNN~\cite{wang2022functionality, wang2024fgnn2}, have introduced pre-training methods that achieve notable results in circuit representation, they primarily focus on structural and functional aspects of AND-Inverter graphs, overlooking other cell types and electrical properties. Moreover, they embed circuit knowledge within the weights of graph neural networks, restricting the transferability of this knowledge to other ML models. Another related research direction focuses on ML-based library cell characterization~\cite{raslan2023deep, hyun2023accurate}. While these methods have shown promise, they primarily aim to improve arc-based timing characterization accuracy rather than enabling ML models to capture and understand semantic relationships among cells. To our knowledge, efficient learning of functional and electrical representations of library cells in a vector space -- compatible with diverse ML architectures (including transformer, the most popular architecture for foundation models) -- remains unexplored. This work addresses that gap by enabling more general and versatile cell representation learning.

\subsection{Challenges}
We identify three key challenges in learning meaningful vector representations of library cells:
\begin{enumerate}
    \item \textbf{Defining and Evaluating Cell Representations}: Defining the semantics of library cells, and devising effective test sets and metrics to assess representation quality.

    \item \textbf{Efficient Training Data Generation}: Creating comprehensive training data that captures the diverse functional and electrical properties of cells.

    \item \textbf{Flexible Model Architecture}: Developing an architecture that accommodates varying pin counts and enables straightforward generation of property-specific (e.g., cell delay-focused) and arc-specific representations from a cell’s base representation. This is critical for downstream circuit learning tasks.
\end{enumerate}










\subsection{Contributions}
Our key insight is that a cell’s semantics are defined by its responses to input conditions, allowing us to capture semantic relationships among cells based on their behaviors under the same inputs. By systematically collecting (input conditions, output responses) pairs, we can create data that encapsulates both the functional and electrical properties of cells. Building on this, we introduce \textit{Lib2Vec}, a novel self-supervised learning framework for library cell representation. Our main contributions are summarized as follows:
\begin{enumerate}
    \item First systematic exploration of learning library cell representation in vector space, adaptable to various ML architectures and circuit applications;
    
    \item An automated method for generating regularity tests to quantitatively evaluate the quality of cell representations;
    
    \item A self-supervised learning scheme that extracts training data from Liberty files, removing the need for costly labeling;

    \item An attention-based architecture that accommodates various pin counts and enables property- and arc-specific embeddings;
    
    \item Experimental results show that Lib2Vec effectively captures functional and electrical properties of cells. Lib2Vec also enhances downstream circuit learning applications, particularly in scenarios with limited labeled data.
\end{enumerate}

The remainder of this paper is organized as follows. \cref{sec:property} introduces key properties of library cells, providing essential background for this study.
\cref{sec:overview} gives an overview of the Lib2Vec framework, with detailed descriptions of its three components in~\cref{sec:regularity_test}, \cref{sec:self_supervised} and \cref{sec:model}. \cref{sec:usage} discusses the usage model of Lib2Vec. Experiment results are shown in \cref{sec:experiment}, and \cref{sec:conclusion} gives some concluding remarks.

%% file: tex/Preliminaries.tex

\begin{figure*}[!t]
\centering
\vspace{-1mm}
\includegraphics[width=\linewidth]{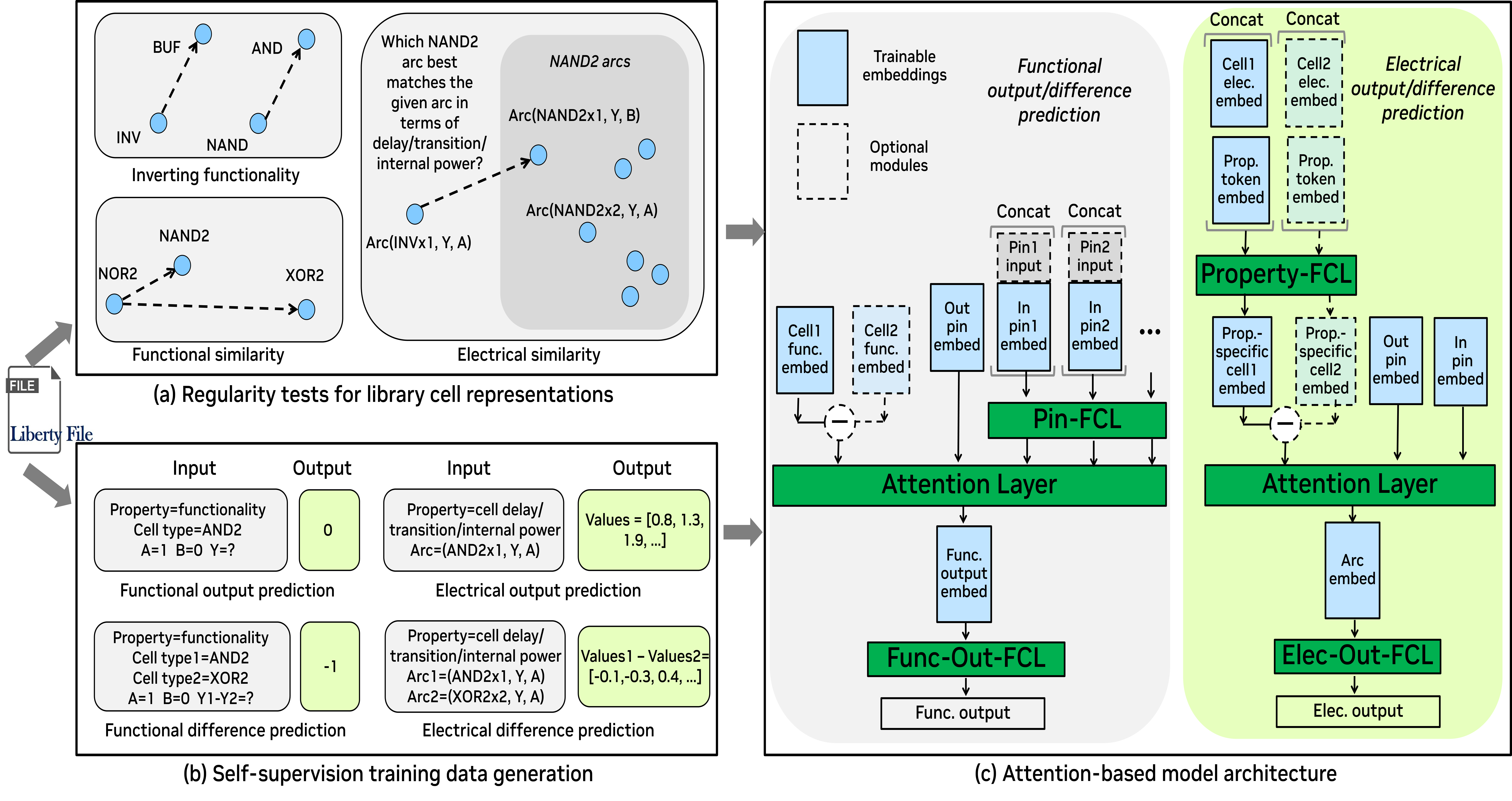}
\vspace{-2mm}
    \caption{Overview of the lib2vec framework.}
    \label{fig:overview}
    \vspace{-3mm}
\end{figure*}

\section{Library Cell Properties}
\label{sec:property}
In VLSI design, library cells have three categories of properties: functional, electrical, and physical. Functional properties define a cell's logical behavior. Electrical properties capture timing, power, and signal integrity, including parameters like propagation delay, transition time, capacitance, leakage and internal power, and noise margins. Physical properties describe a cell’s layout and geometry, such as cell dimensions and pin locations. These properties can be further classified as static or dynamic. Static properties, like physical characteristics, remain constant, while dynamic properties, such as most functional and electrical behaviors, vary with input conditions. This work focuses on dynamic properties, specifically functional and electrical characteristics, as static properties are typically straightforward to model. Specifically, Lib2Vec models functional properties and rise/fall propagation delay, transition time, and internal power. The framework is extensible to incorporate additional properties.

Liberty files~\cite{libertyformat} are a standard format to describe the functional and electrical properties of library cells. In this work, we utilize Liberty files from the ASAP7 cell library~\cite{clark2016asap7} as our test case. A cell's function is described by its functional expression; for instance, the function expression for \textit{AND2x2\_ASAP7\_75t\_R} is \textit{A * B}. In the ASAP7 library, cell propagation delay, transition time, and internal power are characterized as functions of input transition time and total output capacitance, represented through lookup tables. Importantly, the Lib2Vec framework is adaptable to more advanced delay and power models, as long as the output responses of a cell can be efficiently sampled.

%% file: tex/overview.tex
\section{Lib2Vec Overview}
\label{sec:overview}


\cref{fig:overview} depicts the Lib2Vec framework, comprising three key components: (a) an automated method for generating regularity tests, (b) a self-supervised training scheme, and (c) a flexible attention-based model architecture. Self-supervised training data and regularity tests, both derived from Liberty files automatically, remove the need for costly labeling and can be easily adapted to new cell libraries. The attention-based neural networks leverage the generated data to learn cell representations, which are subsequently validated using the regularity tests. Detailed descriptions of each component are provided in the following sections.



%% file: tex/reguality_tests.tex
\section{Definition and Test Sets for Cell Semantics}
\label{sec:regularity_test}
This section tackles the challenge of defining and evaluating cell representations by formally framing the cell representation problem and introducing multiple sets of regularity tests to efficiently assess representation quality.

We propose that a cell's semantics are fully characterized by its responses to specific inputs. Functional and electrical similarities between cells can thus be defined by differences in output responses under identical input conditions. Such similarities are crucial for ML models to analyze and optimize circuit netlist performance while enabling effective cross-cell knowledge transfer. Beyond similarity, functional inversion is another key relationship for tasks like logic propagation and netlist rewriting. Based on these observations, we define three sets of regularity tests that can be automatically derived from Liberty files. The cell representation learning problem can then be formulated as learning vector space representations that maximize accuracy on these regularity tests. Our approach assumes well-documented Liberty files with consistent pin naming, as seen in the ASAP library used in this study. Consequently, input pin reordering is not considered in the regularity tests. Details of the regularity tests are elaborated as follows.

\subsection{Inverting Functionality Tests}
This test set evaluates inverting functionality relationships among cell types. A cell type refers to a group of standard cells with the same functionality but differing in driving strengths, voltage thresholds, or layout implementations. Two cell types with identical input pin names are considered to have an inverting functionality relationship if their outputs always complement each other, such as BUF (buffer) and INV (inverter).

After identifying all inverting functionality pairs, we design tests to evaluate these relationships. For instance, as shown in \cref{fig:overview}(a), given two pairs, \textit{(BUF, INV)} and \textit{(AND2, NAND2)}, two tests are created:

\textit{(BUF vs. INV) = (AND2 vs. ?)}

\textit{(AND2 vs. NAND2) = (BUF vs. ?)}.

More examples can be found in \cref{tab:tests}.
Using linear algebraic operations on cell vectors, we assess whether the target vector (e.g., vector(NAND2)) ranks among the top-K closest vectors to the inferred vector (e.g., vector(INV) - vector(BUF) + vector(AND2)). The resulting top-K accuracy indicates how well the learned cell representations capture inverting functionality relationships.

\subsection{Functional Similarity Tests}
This test set evaluates functional similarity among cell types with identical input pins. To simplify the analysis, we focus on single-output cells, which constitute the majority in the ASAP7 library used in our experiments. Future work will extend functional similarity evaluation to individual output pins.

\begin{figure}[!ht]
\centering
\vspace{-1mm}
\includegraphics[width=\linewidth]{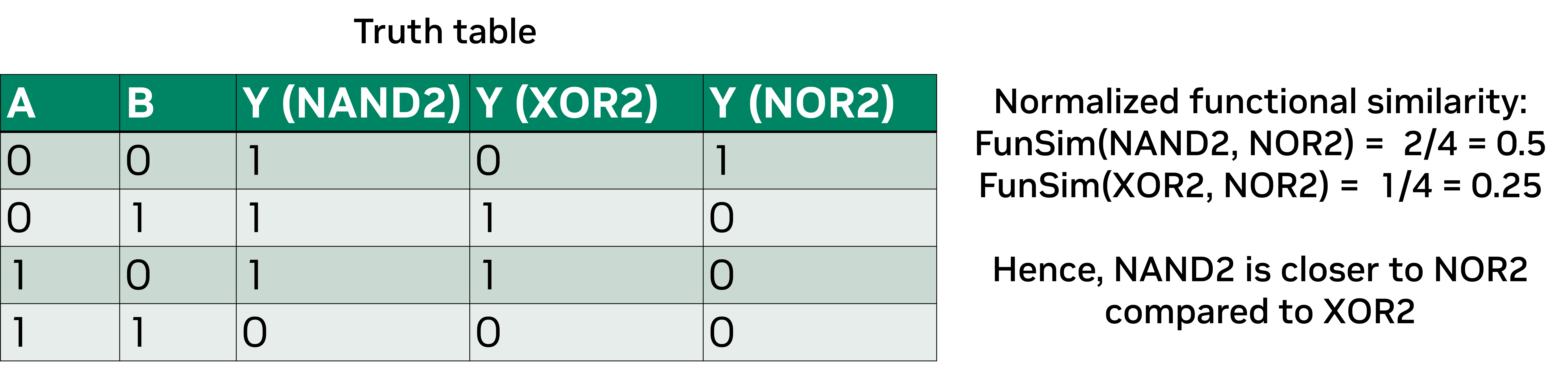}
\vspace{-2mm}
    \caption{Functional similarity calculation for cells with identical input pins.}
    \label{fig:func_sim}
    \vspace{-3mm}
\end{figure}

Functional similarity between two cells is computed by comparing their truth tables, as shown in \cref{fig:func_sim}. It is defined as the ratio of matching output values to the total number of input combinations. For example, the functional similarity between \textit{NAND2} and \textit{NOR2} is \textit{FunSim(NAND2, NOR2)} = $\frac{2}{4}$, while \textit{FunSim(XOR2,NOR2)} = $\frac{1}{4}$. A functional similarity test is created as

\textit{Which is closer to NOR2: NAND2 or XOR2}? And the answer is \textit{NAND2} as \textit{FunSim(NAND2, NOR2)} $>$ \textit{FunSim(XOR2,NOR2)}. 

Functional similarity tests (e.g., which is closer to \textit{C}: \textit{A} or \textit{B}?) are further categorized based on the similarity difference:

Easy test if $0.5 \leq |FunSim(B,C) - FunSim(A,C)|$, the difference is substantial, making the test easier;

Hard test if $0 < |FunSim(B,C) - FunSim(A,C)| < 0.5$, the similarity scores are closer, making the test more challenging.

These tests are answered by comparing the Euclidean distances between functional cell vectors. As binary classification tasks, random guessing yields an accuracy of $50\%$. Higher accuracy indicates that the learned representations effectively capture functional similarity.


\subsection{Electrical Similarity Tests}
This test set evaluates electrical similarity among cell arcs, encompassing rise/fall delay, transition time and internal power. We detail the process of deriving electrical similarity tests using rise delay as an example, which consists of the following $4$ steps:


(1) Input condition sampling: The Non-Linear Delay Model in ASAP7 represents delays using lookup tables parameterized by input slew and output load. To construct input condition combinations, we first determine the maximal ranges of input slew and output load across all cells. After applying a logarithmic transformation to these ranges, we uniformly sample 150 points from each range. This results in $150 \times 150 = 22,500$ input condition combinations, denoted as $\text{\textit{conditions}} = \left[ \left( \text{slew}_1, \text{load}_1 \right), \left( \text{slew}_2, \text{load}_2 \right), \dots \right]$.

(2) Output value calculation: We apply the identical input conditions \text{\textit{conditions}} to all cell arcs and compute the rise delay values. These delay values are then logarithmically transformed to ensure that the distribution approximates a Gaussian distribution, producing a set of transformed delays $\text{\textit{log-delay}} = \left[  \text{log(delay)}_1, \text{log(delay)}_2,  \dots \right]$.

(3) Similarity measurement. We use Euclidean distance to measure the similarity between \text{\textit{log-delay}} vectors of two arcs.

(4) Similarity test creation and evaluation metrics. One example test is: "Which \textit{NOR2} arc is closest to \textit{arc(INVx1,Y,A)} in terms of rise cell delay?" To answer this, we calculate the distance between the rise delay-specific representation of \textit{arc(INVx1,Y,A)} and all arcs in NOR2 cells, then report the \textit{NOR2} arc with the smallest distance.
The top-K accuracy measures how effectively the learned cell representations capture delay-specific similarity relationships.

It is important to note that the test sets described above are not exhaustive in defining what Lib2Vec can capture. They are designed for fast evaluation of the quality of cell representations, but they do not constrain the scope of what Lib2Vec can learn. For instance, as illustrated in \cref{fig:visualization} (b), Lib2Vec identifies functional similarities between cells such as \textit{AO222}, \textit{AO322}, and \textit{AO332}, despite their differing input configurations.

%% file: tex/self_supervision.tex
\section{Library Cell Self-Supervised Learning Scheme}
\label{sec:self_supervised}
This section addresses the challenge of training data generation for cell representations. An automatic method is proposed to create comprehensive functional and electrical data from Liberty files. 

In natural language processing, masked prediction -- predicting missing words based on context~\cite{mikolov2013efficient} -- has proven effective for generating word representations, as a word's semantics are defined by its context. Inspired by this, we propose novel self-supervised learning methods tailored to capture the semantics of cells. Since a cell's semantics are determined by its response to input conditions, we introduce four self-supervised tasks, where training data is derived from the functional and electrical responses of cells, as shown in \cref{fig:overview} (b):



(1) Functional output prediction: It predicts the output logic value of a cell given its input logic values. \textit{Example:} For an \textit{AND2} cell, with inputs \textit{A}=1 and \textit{B}=0, the output \textit{Y} is ? (answer: 0).

(2) Functional difference prediction: It predicts the output logic value difference between two cells given the same input logic values.  
    \textit{Example:} For cells \textit{AND2} and \textit{XOR2}, with inputs \textit{A}=1 and \textit{B}=0, the difference \textit{Y(AND2)} - \textit{Y(XOR2)} is ? (answer: -1).

(3) Electrical output prediction: it predicts the electrical property values (e.g., delay, power, transition) of a specific cell arc under the input conditions \text{\textit{conditions}} introduced in \cref{sec:regularity_test}-C.  
    \textit{Example:} For the \textit{arc(AND2x1,Y,A)}, the rise cell delay is ? (answer:\([0.8, 1.3, 1.4, \dots]\)).

(4) Electrical difference prediction: It predicts the difference in electrical property values between two cell arcs under the same \text{\textit{conditions}}.  
    \textit{Example:} For \textit{arc(AND2x1,Y,A)} and \textit{arc(XOR2x2,Y,A)}, the rise cell delay difference is ? (answer:\([-0.1, -0.3, 0.2, \dots]\)).





Difference prediction data emphasizes how cells differ in functionality or electrical properties, complementing to the absolute output value prediction. These tests ensure the model captures subtle relationships between cells, improving robustness and aligning with real-world design tasks that rely on comparing cell behaviors. 











%% file: tex/model.tex
\section{Model Architecture}
\label{sec:model}
This section presents an attention-based model architecture designed to efficiently process functional and electrical datasets introduced in \cref{sec:self_supervised}. The architecture ensures consistent-length vector representations for cells with different input/output configurations, while also supporting property-specific representations for both entire cells and individual timing arcs.

Since a cell's functional properties are independent of its electrical properties, two separate models are developed to learn functional and electrical representations. Despite being distinct, the two models share a similar architecture, as shown in \cref{fig:overview}(c). The proposed architecture includes learnable embeddings for cells, pin names (shared across cells), and properties (e.g., rise delay). For functional output prediction, the attention layer generates the functional embedding for an output pin by attending to the cell's functional embedding and the embeddings of all corresponding pins. This attention mechanism allows the model to accommodate cells with varying pin counts. Multiple fully connected layers, referred to as \textit{Func-Out-FCL} in \cref{fig:overview}(c), then transform the functional embedding of the output pin into a logic value prediction.

For electrical output prediction, an electrical property-specific (e.g., rise delay) cell representation is created by concatenating the base electrical embedding of the cell with the property token embedding and passing them through the fully connected layer \textit{Property-FCL}. 
Since the same input conditions are applied to all arcs, we do not take the input conditions as input. An attention layer then combines the property-specific cell embedding with the input and output pin embeddings to create the timing arc embedding. The \textit{Elec-Out-FCL} further maps this arc embedding to the electrical output prediction.

For functional and electrical difference prediction tasks, the model includes an additional branch to compute the embeddings and differences between two cells, as depicted by the optional modules in \cref{fig:overview}(c). This architecture offers flexibility to adapt to various learning tasks.

To encourage the models to encode cell knowledge within the cell embeddings rather than the weights of the attention and fully connected layers, we limit the number of learnable parameters in these layers. Specifically, we use a single-head attention operator in the \textit{Attention Layer} module and two-layer fully connected operators in the \textit{FCL} modules.

%% file: tex/usage.tex
\section{Utilizing Lib2Vec for Downstream Applications}
\label{sec:usage}
We propose two strategies for integrating Lib2Vec into ML models for downstream applications:

(1) Representation-based integration: Directly use pre-trained cell embeddings or property-specific cell/arc representations as input features for downstream tasks. This approach is simple and compatible with a wide range of ML models.

(2) Model-based integration: Incorporate the proposed attention-based architecture into downstream ML models for circuit applications. By using Lib2Vec's self-supervised training as a pretraining step, both cell embeddings and model weights are initialized effectively. This tightly integrates Lib2Vec with the downstream task, potentially yielding greater performance benefits compared to the first approach.

%% file: tex/Experiment.tex
\section{Experimental Results}
\label{sec:experiment}
\subsection{Lib2Vec Implementation Details}
The Lib2Vec framework was implemented in Python. A custom Liberty parser for the ASAP7 cell library was developed based on \cite{liberty}. Regularity tests and self-supervised training datasets were generated using tailored Python scripts. The ASAP7 library contains $190$ standard cells, which can be grouped into $86$ cell types according to their functional expressions. The attention-based models were implemented in PyTorch and trained on a Linux machine equipped with an AMD EPYC 7742 64-Core Processor and Nvidia A100 GPUs. We explored various embedding sizes to assess their impact on capturing cell relationships. The runtime for training data generation is approximately $10$ minutes. Learning functional representations takes around $20$ minutes, while learning electrical representations requires about $4$ hours on one GPU.


\subsection{Regularity Test Results}
\cref{tab:tests} presents examples of regularity tests and their evaluation metrics, encompassing various functional and electrical relationships among cells. In total, $930$ inverting functionality tests are created. Regarding functional similarity, $116$ easy and $166$ hard tests are generated. In terms of electrical similarity, $635$, $467$, $975$, $858$, $722$ and $722$ tests are created for rise delay, fall delay, rise transition, fall transition, rise internal power and fall internal power, respectively. Answering these tests requires both functional and electrical cell representations, as well as property-specific (delay/transition/power) arc representations. To the best of our knowledge, no existing methods offer this level of flexibility. Therefore, we compare Lib2Vec's performance against random guessing.

\begin{figure}[!t]
\centering
\vspace{-1mm}
\includegraphics[width=1.0\linewidth]{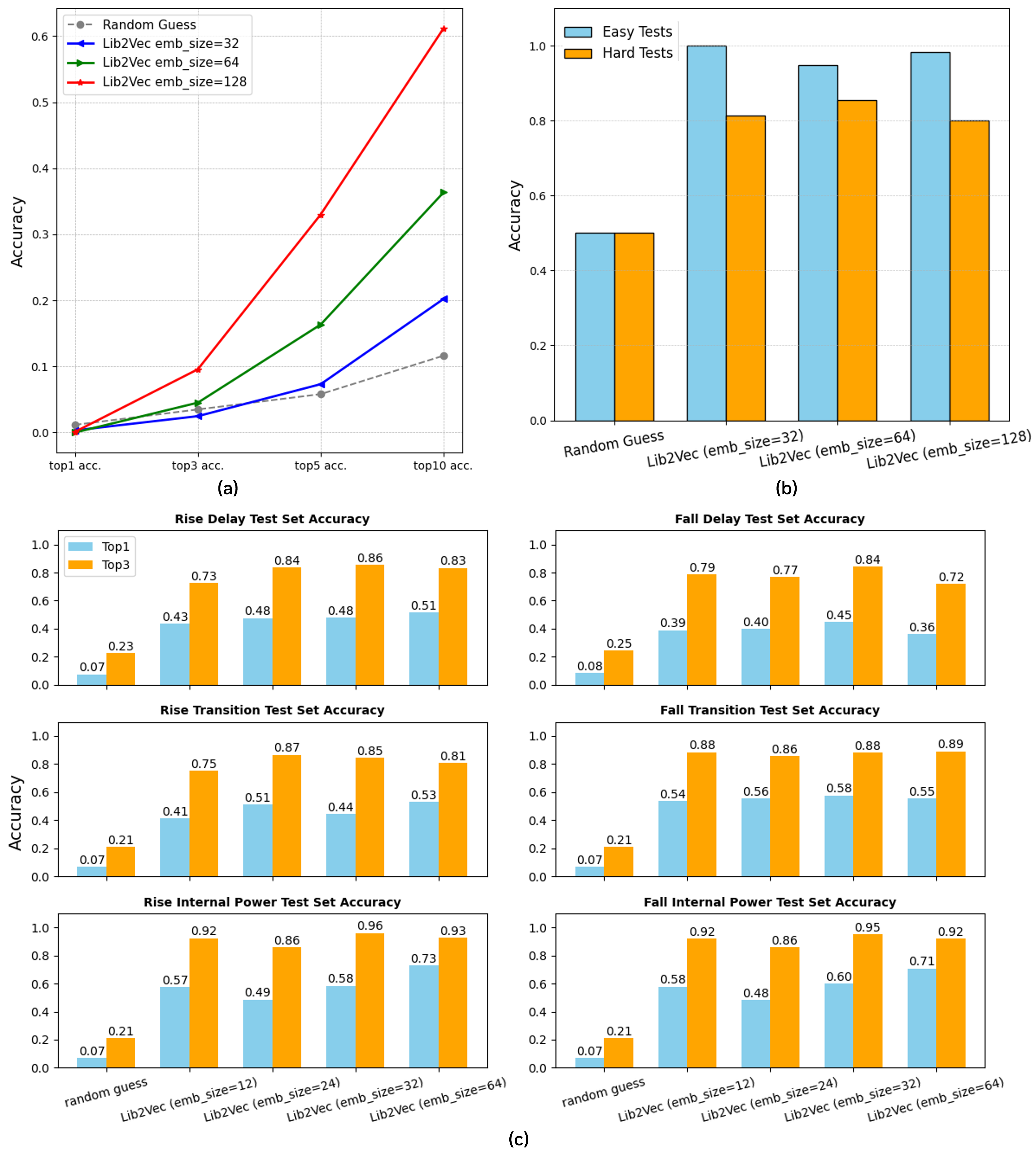}
\vspace{-2mm}
    \caption{Accuracy comparison in (a) inverting functionality, (b) functional similarity and (c) electrical similarity test sets between random guess and Lib2Vec with various embedding sizes.}
    \label{fig:function_test}
    \vspace{-3mm}
\end{figure}

\cref{fig:function_test}(a), (b), and (c) show the results on the inverting functionality, functional similarity, and electrical similarity test sets, respectively. Lib2Vec consistently outperforms random guessing. For inverting functionality (\cref{fig:function_test}(a)), Lib2Vec with larger embedding size leads to higher accuracy. And Lib2Vec with an embedding size of 64 achieves a top-10 accuracy of $61\%$, compared to $11\%$ for random guess. For functional similarity (\cref{fig:function_test}(b)), different embedding sizes result in similar performance. Lib2Vec achieves near-perfect accuracy on easy tests and over $80\%$ on hard tests, far surpassing the $50\%$ baseline. For electrical similarity (\cref{fig:function_test}(c)), Lib2Vec excels across all metrics. On average, Lib2Vec with embedding size $32$ achieves $52\%$ top-1 accuracy and $89\%$ top-3 accuracy in electrical similarity tests, significantly outperforming $7\%$ top-1 and $22\%$ top-3 accuracies of random guessing. These results validate Lib2Vec's ability to capture both functional and electrical properties. 

Additional ablation studies were conducted to evaluate the impact of functional/electrical \textit{difference} prediction tasks and of key model architecture decisions. We find that excluding the difference prediction datasets destabilizes Lib2Vec training and reduces accuracy in the regularity tests. Furthermore, replacing the single-layer, single-head attention operator in our model with a more complex two-layer, two-head attention mechanism lowers training loss but degrades regularity test accuracy. This suggests that limiting the number of trainable parameters in the attention operator encourages the model to encode cell knowledge effectively into the representations.

\begin{table*}[!t]
\vspace{-4mm}%
\caption{Examples of regularity tests and evaluation metrics}
\label{tab:tests}
\vspace{-1mm}%
\centering
\scriptsize
\begin{tabularx}{\linewidth}{|c|c|X|c|X|}
\hline
Relationship                     &                                      & Question                                                          & Answer               & Evaluation metrics                                                                                                                                                                                                                                                             \\ \hline \hline
\multirow{5}{*}{\makecell[l] {Inverting\\ functionality}} & \multirow{5}{*}{}                    & (BUF vs. INV) =   (AND2 vs. ?)                                    & NAND2                & \multirow{5}{*}{\makecell[l] {Use linear algebraic operations on cell vectors to\\determine the answer. E.g., assess whether\\ vector(NAND2) falls within the top-K closest\\vectors to vector(INV)-vector(BUF)+ vector(AND2),\\ and report the resulting top-K accuracy}} \\ \cline{3-4}
                                         &                                      & (BUF vs. INV) = (XNOR2 vs. ?)                                     & XOR2                 &                                                                                                                                                                                                                                                                                \\ \cline{3-4}
                                         &                                      & (AO211 vs. AOI211) = (OR2 vs. ?)                                  & NOR2                 &                                                                                                                                                                                                                                                                                \\ \cline{3-4}
                                         &                                      & (OR5 vs. NOR5) = (OA333 vs. ?)                                    & OAI333               &                                                                                                                                                                                                                                                                                \\ \cline{3-4}
                                         &                                      & (MAJ vs. MAJI) = (AND5 vs. ?)                                     & NAND5                &                                                                                                                                                                                                                                                                                \\ \hline \hline
\multirow{6}{*}{\makecell[l] {Functional\\ similarity}}   & \multirow{3}{*}{Easy}                & 
Which is closer to AO21: OA21 or AOI21?                     & OA21                 & \multirow{6}{*}{\makecell[l] {Determine the answer by evaluating the Euclidean\\distance between functional cell vectors,\\ and report the accuracy of the binary classification}}                                                                                                               \\ \cline{3-4}
                                         &                                      & Which is closer to NAND5: OR5 or NOR5?                      & OR5                  &                                                                                                                                                                                                                                                                                \\ \cline{3-4}
                                         &                                      & Which is closer to NOR4: AND4 or NAND4?                     & AND4                 &                                                                                                                                                                                                                                                                                \\ \cline{2-4}
                                         & \multirow{3}{*}{Hard}                & Which is closer to A2O1A1I: O2A1O1I or AO211?               & O2A1O1I              &                                                                                                                                                                                                                                                                                \\ \cline{3-4}
                                         &                                      & Which is closer to A2O1A1I: OAI211 or AOI211?               & OAI211               &                                                                                                                                                                                                                                                                                \\ \cline{3-4}
                                         &                                      & Which is closer to NOR2: NAND2 or XOR2?                     & NAND2                &                                                                                                                                                                                                                                                                                \\ \hline \hline
\multirow{12}{*}{\makecell[l] {Electrical\\ similarity}}  & \multirow{2}{*}{Rise delay}          & Which NOR2 arc is closest to arc(INVx1, Y, A)       & arc(NOR2x1,Y,B)    & \multirow{12}{*}{\makecell[l] {Determine the answer by evaluating the Euclidean\\ distance between delay/transition/power-specific\\ cell arc vectors and report top-K accuracy}}                                                                                                                                                                                            \\ \cline{3-4}
                                         &                                      & Which NAND2 arc is closest to arc(INVxp33,Y,A)     & arc(NAND2xp33,Y,B) &                                                                                                                                                                                                                                                                                \\ \cline{2-4}
                                         & \multirow{2}{*}{Fall delay}          & Which NOR2 arc is closest to arc(A2O1A1Ixp33,Y,A1) & arc(NOR2xp67,Y,A)  &                                                                                                                                                                                                                                                                                \\ \cline{3-4}
                                         &                                      & Which BUF arc is closest to arc(AO211x2,Y,A1)        & arc(BUFx8,Y,A)       &                                                                                                                                                                                                                                                                                \\ \cline{2-4}
                                         & \multirow{2}{*}{\makecell[l]{Rise \\transition}}     & Which NAND2 arc is closest to arc(INVx1,Y,A)         & arc(NAND2x1,Y,B)     &                                                                                                                                                                                                                                                                                \\ \cline{3-4}
                                         &                                      & Which NAND2 arc is closest to arc(INVx2,Y,A)         & arc(NAND2x2,Y,B)     &                                                                                                                                                                                                                                                                                \\ \cline{2-4}
                                         & \multirow{2}{*}{\makecell[l]{Fall \\transition}}     & Which BUF arc is closest to arc(AO211x2,Y,A1)        & arc(BUFx2,Y,A)       &                                                                                                                                                                                                                                                                                \\ \cline{3-4}
                                         &                                      & Which BUF ar is closest to arc(AO211x2,Y,A2)        & arc(BUFx4,Y,A)       &                                                                                                                                                                                                                                                                                \\ \cline{2-4}
                                         & \multirow{2}{*}{\makecell[l] {Rise internal\\ power}} & Which NOR2 arc is closest to arc(INVx1,Y,A)          & arc(NOR2x1,Y,A)      &                                                                                                                                                                                                                                                                                \\ \cline{3-4}
                                         &                                      & Which NOR2 arc is closest to arc(INVx2,Y,A)          & arc(NOR2x2,Y,A)      &                                                                                                                                                                                                                                                                                \\ \cline{2-4}
                                         & \multirow{2}{*}{\makecell[l] {Fall internal\\ power}} & Which BUF arc is closest to arc(AO211x2,Y,A1)        & arc(BUFx2,Y,A)       &                                                                                                                                                                                                                                                                                \\ \cline{3-4}
                                         &                                      & Which BUF arc is closest to arc(AO211x2,Y,A2)        & arc(BUFx2,Y,A)       &                                                                                                                                                                                                                                                                                \\ \hline
\end{tabularx}
\end{table*}

\subsection{Visualization of the Cell Representations}
We visualize cell representations learned through masked prediction and Lib2Vec using the t-SNE technique~\cite{van2008visualizing}. For masked prediction, we collect approximately $30,000$ timing paths from $10$ post-route designs in the IWLS 2005 benchmark suite~\cite{albrecht2005iwls}, synthesized with commercial EDA tools at the ASAP7 technology node. Cell representations are learned using a TransSizer~\cite{nath2022transsizer}-style transformer model with an embedding size of $32$, trained to predict masked cells within timing paths.

\begin{figure*}[!t]
\centering
\vspace{-1mm}
\includegraphics[width=1.0\linewidth]{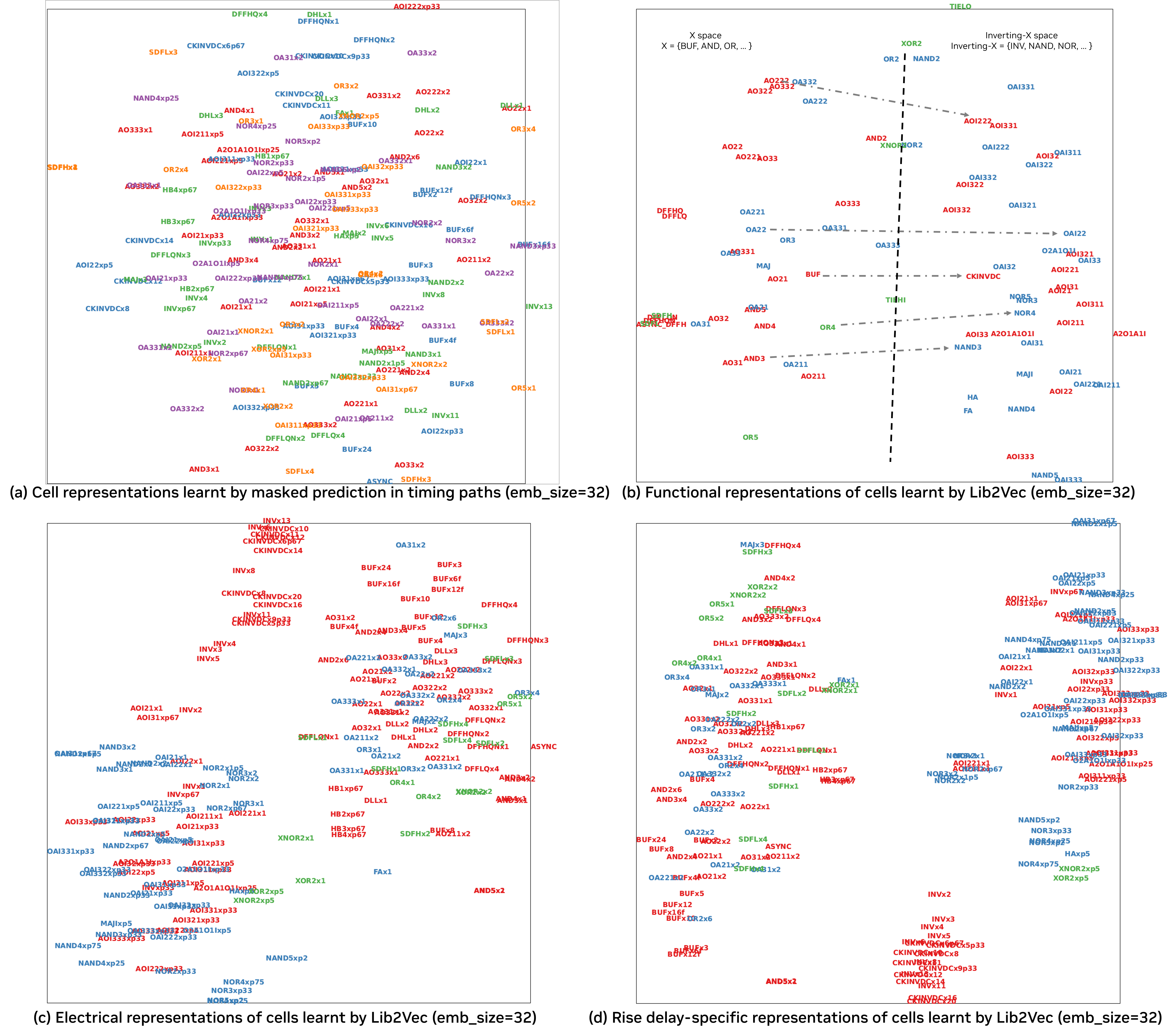}
\vspace{-1mm}
    \caption{Visualization of cell representations.}
    \label{fig:visualization}
\end{figure*}

As shown in \cref{fig:visualization}, the cell representations learned by Lib2Vec reveal clear, intuitive relationships, while the representations learned through masked prediction do not exhibit discernible patterns. The functional embedding space produced by Lib2Vec effectively captures the complex functional relationships among cells. Notably, the space is naturally divided into a \textit{X} space (\textit{X={BUF,AND,OR,...}}) and an inverting-\textit{X} space (inverting-\textit{X={INV,NAND,NOR,...}}). Linear algebraic operations on the representations reveal interesting relationships, such as  \textit{vector(BUF) - vector(INV)} $\approx$ \textit{vector(NAND3) - vector(AND3)} $\approx$ \textit{vector(NOR4) - vector(OR4)}, depicted in \cref{fig:visualization}(b). \cref{fig:visualization}(d) visualizes the rise delay-specific cell representations learned by Lib2Vec. Using \textit{INV} cells as an example, the cell representations space clearly captures the driving strengths ordering of \textit{INV} cells.



\subsection{Results on Integrating Lib2Vec for Downstream Tasks}
We integrate Lib2Vec into a graph neural model for three netlist logic prediction tasks. Given a netlist represented as a directed acyclic graph and input vectors for its input ports, the model predicts (1) the output vector at each cell's output pin, (2) the logic probability (probability of logic `1') and (3) switching activity. Achieving high accuracy on these tasks requires the model to effectively approximate logic functions of various cell types, manage propagation across the graph, and mitigate error accumulation. To ensure diverse datasets and control netlist dimensions, we develop an artificial netlist generator to create $1,351$ netlists using cells from the ASAP library. The statistics of these netlists are detailed in \cref{tab:netlists}. 

A custom graph attention network is developed to perform attention-based message passing in topological order, approximating logic propagation within a netlist. Node features are cell representations, while edge features combine the driver and sink pin representations. Both cell and pin representations are learnable vectors of length $32$. In the baseline random initialization, these representations and network weights are initialized randomly. It essentially employs one-hot encoding for cell/pin representations. For representation-based Lib2Vec integration, functional embeddings from Lib2Vec serve as cell/pin representations, with other parameters initialized randomly. In model-based integration, cell/pin features and model weights (especially the attention operator parameters for message passing) are pretrained using Lib2Vec's self-supervised framework. The primary goal is to assess whether Lib2Vec enhances learning in scenarios with limited labeled data. To ensure a fair comparison, consistent training and testing data splits, as well as identical hyperparameters, are used across all methods, with results averaged over three runs. Each training run requires about one day on a GPU, while the Lib2Vec pretraining process completes in about 15 minutes. 
The proportion of training data varies from as low as $0.3\%$ ($4$ samples) to $10\%$ ($135$ samples) and performance is evaluated on the remaining dataset.

\begin{figure*}[!t]
\centering
\includegraphics[width=1.0\linewidth]{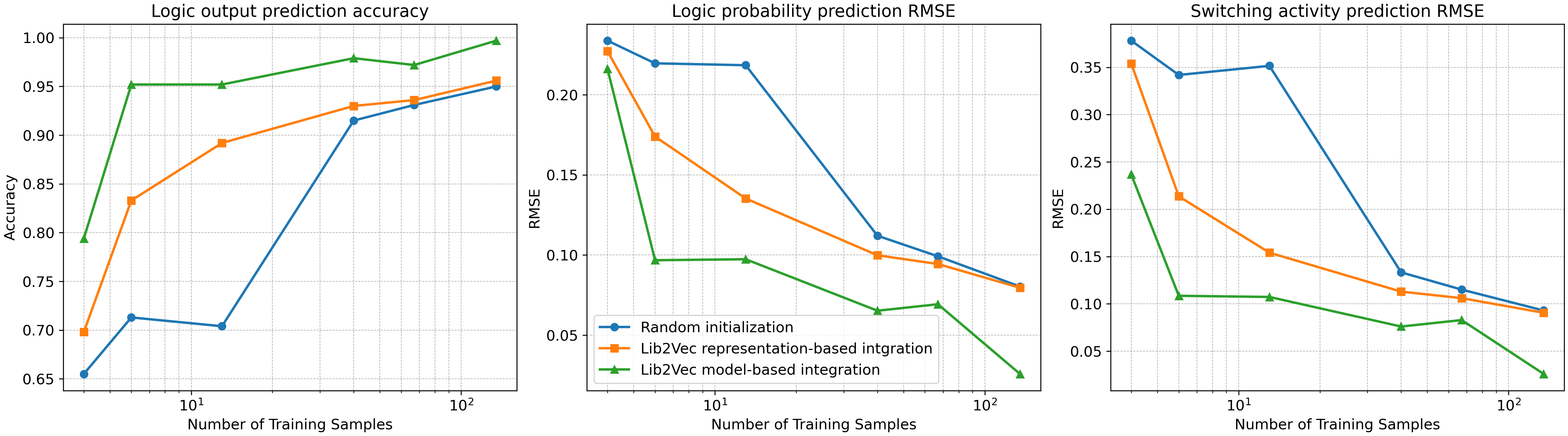}
    \caption{Impacts of integrating Lib2Vec into ML models for different prediction tasks.}
    \label{fig:integration}
\end{figure*}

As shown in \cref{fig:integration}, both integration methods outperform random initialization, particularly in low-data regimes, demonstrating that Lib2Vec effectively transfers cell knowledge to aid circuit learning. Notably, the model-based integration achieves approximately $80\%$ accuracy in logic output prediction and $0.236$ RMSE (Root Mean Square Error) with only $4$ training samples, comparing to the $65\%$ accuracy and $0.378$ RMSE of random initialization. It highlights Lib2Vec's potential for enabling few-shot learning. Further investigation is needed to fully understand and expand this capability.



\begin{table}[!t]
\caption{Statistics of generated netlists}
\label{tab:netlists}
\begin{tabular}{|l||l|l|l|l|}
\hline
      & \#cells       & \#input ports & \#edges       & \#topological levels \\ \hline \hline
range & {[}16, 235{]} & {[}1, 16{]}   & {[}34, 875{]} & {[}7, 111{]}    \\ \hline
mean  & 117           & 10            & 421           & 51              \\ \hline
\end{tabular}
\end{table}


%% file: tex/conclusion.tex
\section{Conclusion}
\label{sec:conclusion}
In this paper, we introduce Lib2Vec, a novel self-supervised framework for learning meaningful vector representations of library cells, enabling ML models to capture functional and electrical relationships. Future work includes exploring Lib2Vec for circuit foundation models and systematically evaluating its impact on downstream tasks.